\documentclass[runningheads]{llncs}
\usepackage{graphicx}
\usepackage{cite}
\usepackage{amsmath,amssymb,amsfonts}
\usepackage{graphicx}
\usepackage{textcomp}
\newcommand{\acknowledgmentsname}{Acknowledgments}
\usepackage{algorithm}
\usepackage{algpseudocode}
\usepackage[table,xcdraw]{xcolor}
\begin{document}
\title{Learning from Peers: Collaborative Ensemble Adversarial Training}
%
%

\author{\textsuperscript{1}Dengjin Li, \textsuperscript{1}Yanming Guo, Yuxiang Xie, Zheng Li, \\ Jiangming Chen, Xiaolong Li, Mingrui Lao \thanks{Corresponding Author: Mingrui Lao}}

\institute{Laboratory of Big Data and Decision, National University of Defense Technology, Changsha, Hunan, China \\
\{lidengjin, guoyanming, yxxie, lizheng18, lixiaolong\}@nudt.edu.cn, jiangming\_chen@126.com, laomingrui@vip.sina.cn
}

\maketitle         
\begin{abstract}
Ensemble Adversarial Training (EAT) attempts to enhance the robustness of models against adversarial attacks by leveraging multiple models. However, current EAT strategies tend to train the sub-models independently, ignoring the cooperative benefits between sub-models. Through detailed inspections of the process of EAT, we find that that samples with classification disparities between sub-models are close to the  decision boundary of ensemble, exerting greater influence on the robustness of ensemble. To this end, we propose a novel yet efficient Collaborative Ensemble Adversarial Training (CEAT), to highlight the cooperative learning among sub-models in the ensemble. To be specific, samples with larger predictive disparities between the sub-models will receive greater attention during the adversarial training of the other sub-models. CEAT leverages the probability disparities to adaptively assign weights to different samples, by incorporating a calibrating distance regularization. Extensive experiments on widely-adopted datasets show that our proposed method achieves the state-of-the-art performance over competitive EAT methods. It is noteworthy that CEAT is model-agnostic, which can be seamlessly adapted into various ensemble methods with flexible applicability.

\keywords{Collaborative Strategy  \and Ensemble Adversarial Training \and Calibrating Regularization.}
\end{abstract}

\section{Introduction}

\label{sec:intro}

Adversarial samples are crafted attacks on models\cite{b1}\cite{b2} that involve adding imperceptible perturbations to natural samples. These perturbations may not be noticeably different from real samples to the human eye, but they can easily cause the model to make incorrect classifications, which has a negative impact on the field of deep learning\cite{b3}. In recent years, defense strategies\cite{b4} have been developed, where adversarial training is emerging as the most effective and practical approach. However, the adversarial training of a single model may not fully address the urgent need for robustness, ensemble adversarial training (EAT) \cite{b5} incorporates multiple sub-models into the training process. Mainstream research in EAT typically concentrates on promoting the predictive diversity\cite{b6}\cite{b7} among sub-models, thereby reducing the transferability of generated adversarial samples\cite{b8}\cite{b9} among sub-models. Nevertheless, these EAT methods still fail to achieve strong robustness as expected\cite{b10}\cite{b11}.

\begin{figure}[t]
    \centering
    \includegraphics[width=7.5cm, height=5cm]{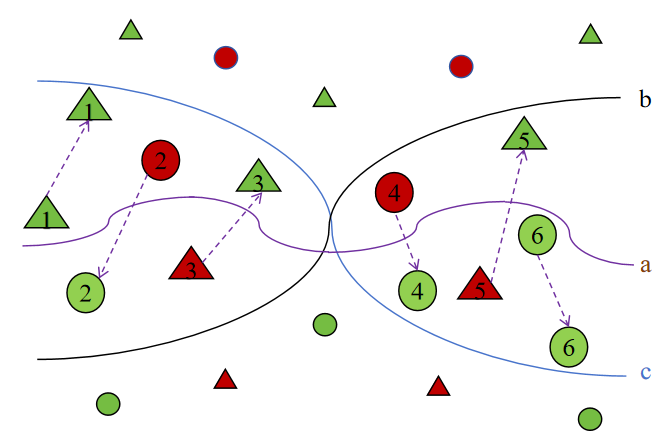}
    \caption{Lines a, b, and c are decision boundaries of sub-models in the ensemble. The green indicates correct classification and the red indicates incorrect classification in ensemble. The purple arrow represents the classification performance after training with collaborative strategy.}
    \label{fig:reweighting}
\end{figure}

Sub-models in most EAT paradigms \cite{b12}\cite{b13} are trained independently, without fully leveraging the interactions among predictions from different models. It would impair the discriminative ability of EAT for training samples with different adversarial robustness. For example, we use three sub-models in the ensemble, represented as model a, model b, and model c. In contrast, we assume the aforementioned sample importance for the one sub-model, can be captured from the prediction disparities between other the two sub-models, thereby fulfilling better adversarial robustness. Based on above analysis, this paper proposes an innovative collaborative ensemble adversarial training (CEAT) to train samples with dynamical weights by cooperation between sub-models.

To further clarify the motivation of CEAT, in Figure \ref{fig:reweighting}, we use triangles and circles to represent two classes of samples that are easily misclassified. The samples lying between the decision boundaries of sub-models b and c, mean that one sub-model classifies them correctly while the other classifies them incorrectly. Whether these samples are classified correctly in EAT paradigm depends on the prediction of the sub-model a. In this case, our CEAT leverages the sub-model collaboration, and assign more training importance over these samples when training sub-model a. Thus, CEAT can correct the misclassified samples by effectively pulling them towards the right side of the decision boundary.

The main contributions of this paper as follows:
\begin{itemize}

    \item This paper is the further work to improve EAT robustness through the collaborations between sub-models. 
    It reveals the different sample contributions towards adversarial robustness for the one sub-model, which can be captured from the prediction disparities between the other sub-models.
    
    \item We propose a novel yet efficient Collaborative Ensemble Adversarial Training(CEAT) that involves a dynamic calibrating distance regularization collaboratively based on the predictive disparities between the other sub-models. 
    
    \item Extensive experimental results on three widely-used datasets demonstrate the state-of-the-art performance of our method. In addition, our proposed method can also serve as a plug-and-play module, which can be flexibly integrated into previous methods to improve adversarial robustness.
    
\end{itemize}

\section{Related Work}

Ensemble adversarial training primarily targets two key research directions: promoting diversity in model outputs and minimizing the transferability of adversarial examples across different sub-models. For instance, Adaptive Diversity Promoting (ADP) \cite{b6} introduced a regularization method that encourages variability in non-maximum predictions. Gradient Alignment Loss (GAL) \cite{b7} aimed to minimize the overlap of adversarial subspaces between sub-models. Transferability Reduced Smooth (TRS) \cite{b8} simultaneously promoted gradient orthogonality and smoothed across sub-models. Diversifying Vulnerability for Enhanced Robust Generation of Ensembles (DVERGE) \cite{b7} focused on isolating and extracting non-robust features from each sub-model. LAFED \cite{b13} minimized the similarity between learning representations by integrating vulnerability data from other sub-models in the latent space. FASTEN \cite{b10} enhanced data quality by recovering knowledge from training trajectories and distinguished features belonging to different categories.

\section{METHODOLOGY}

\subsection{Preliminary}

For a K-class classification problem, given dataset$\left\{\left(x_i, y_i\right)\right\}$, where $x_i$ and $y_i$ represent clean samples and corresponding labels, respectively. We can obtain adversarial samples $\widetilde{x}_i$ by adding perturbations to clean samples. Ensemble is the process of averaging the output results of all sub-models. The distribution of prediction probabilities for the ensemble on the adversarial samples is:

\begin{equation}
  f_{en}\left(\widetilde{x}_i\right) = \underset{k=1, \ldots, K}{\arg \max } \hspace{0.1cm}
  \frac{1}{M} \sum_{m=1}^M p_m(\widetilde{x}_i, k),
\end{equation}
where $k$ refers to class, $m$ denotes one sub-model and $p_{m}(\widetilde{x}_i, k)$ is the output probability of the sub-model after applying softmax to the logits, $f_{en}$ is the output of the ensemble, and $M$ is the number of sub-models.

Generating adversarial samples requires multiple iterations to calculate the gradient of the adversarial samples. These iterations generate perturbations for both the single model and the ensemble model, respectively, as: 
 
\begin{equation}
   \delta_m = \alpha * \operatorname{sgn}(\nabla_x {L}(f_m\left(\widetilde{x}_i\right),y_i)),
   \label{eq:single_model}
\end{equation}

\begin{equation}
   \delta_{en} = \alpha * \operatorname{sgn}(\nabla_x {L}\left(f_{en}\left(\widetilde{x}_i\right), y_i\right)), 
   \label{eq:ensemble_model}
\end{equation}
where $\alpha$ denotes the step size, which controls the magnitude of the update at each iteration and $\nabla_x L$ denotes the gradient of the loss function with respect to the input. sgn is the sign function, which is used to determine the direction of the perturbation. The difference between these two types of generated perturbation is that Equation (\ref{eq:single_model}) utilizes gradient information of single model, while Equation (\ref{eq:ensemble_model}) utilizes gradient information of the ensemble.

The basic idea of adversarial training is to input adversarial samples to train the model, making it robust to adversarial attacks. Adversarial training can be expressed as a min-max problem:
\begin{equation}
\mathcal{L}_{\mathrm{AT}} = \underset{\theta}{\min} \hspace{0.1cm} \frac{1}{n} \hspace{0.1cm} \sum_{i=1}^n \underset{\|\delta\|_p \leq \epsilon}{\max} {L}\left({x}_i+\delta_m, y_i; \theta\right),
\end{equation}
where $\theta$ is the parameter of the model, $n$ is the total number of samples and $\epsilon$ is the range of perturbations. Ensemble takes advantage of the complementary strengths between sub-models so that perturbations cannot precisely attack the vulnerabilities of each sub-model. Adversarial samples generated by the perturbations adding to the ensemble motivate sub-models to focus on the vulnerabilities of the ensemble to optimize. Adversarial training process for sub-models in the ensemble as follows:

\begin{equation}
\mathcal{L}_{\mathrm{EAT}} = \underset{\theta}{\min} \hspace{0.1cm} \frac{1}{n} \hspace{0.1cm} \sum_{i=1}^n \underset{\|\delta\|_p \leq \epsilon}{\max} {L}\left({x}_i+\delta_{en}, y_i; \theta\right).
\end{equation}

The internal maximization refers to finding the worst disturbance, that is, within the disturbance range, to maximize the degree of prediction error of the model. The external minimization is adjusting the model parameters to minimize the average loss of the model for all training samples, even under the worst perturbation conditions. Sub-models adopt $\mathcal{L}_{\mathrm{EAT}}$ optimization that can learn adversarial samples with more diverse features to compensate for vulnerabilities of the ensemble. Each sub-model has an independent optimizer that updates parameters based on its own loss. We will choose this optimization as the baseline for subsequent analysis.

\subsection{Samples with classification disparities between sub-models}

Vallina AT generates adversarial samples through the ensemble, which are then fed into sub-models for adversarial training. Each sub-model has an independent optimizer that updates parameters independently according to its specific loss. The training loss for sub-model is as follows:
\begin{equation}
 \mathcal{L}_{ce}=C E\left(f_m\left(\widetilde{x}_i\right), y_i\right).
\label{eq:ce}
\end{equation}

In previous ensemble methods, all training samples are treated equally without particular emphasis on samples that are prone to classification errors. We aim to filter out the samples with prediction disparities between sub-models and give special attention to these samples. Adversarial risk of dataset$\left\{\left(x_i, y_i\right)\right\}$ for sub-models can be defined the 0-1 loss as:
\begin{equation}
R\left(\theta_m\right)=\frac{1}{n} \sum_{i=1}^n \max_{\substack{\widetilde{x}_i \in \mathcal{B}_\epsilon\left(x_i\right)}}
I\left(f_m\left(\widetilde{x}_i\right) \neq y_i\right),
\end{equation}
 where $I$ is the indicator function, which takes the value 1 when the  prediction on samples does not match the true label, indicating the presence of adversarial risk; otherwise, $I$ takes the value 0, indicating that the sample is relatively stable and less susceptible to attacks. If more than half of sub-models classify the samples incorrectly, the ensemble will have high adversarial risk. Adversarial samples can be divided into two subsets based on the prediction results of sub-model: correctly classified samples $S_m^{+}\left(\widetilde{x}_i\right)$ and misclassified samples $S_m^{-}\left(\widetilde{x}_i\right)$:

\begin{equation}
S_m^{+}\left(\widetilde{x}_i\right)=\left\{i: i \in[n], f_m\left(\widetilde{x}_i\right)=y_i\right\},
\end{equation}

\begin{equation}
S_m^{-}\left(\widetilde{x}_i\right)=\left\{i: i \in[n], f_m\left(\widetilde{x}_i\right) \neq y_i\right\}.
\end{equation}

If there are three sub-models in the ensemble, denoted as model a, model b, model c. When training model a, we need to filter samples with inconsistent classification between model b and model c. We define the \textit{Filter Functions}:

\begin{equation}
   \textit{F}_1\left(S_b^{+}, S_c^{-}\right) = \left\{\left(f_b\left(\widetilde{x}_i\right) = y_i\right) \cap \left(f_c\left(\widetilde{x}_i\right) \neq y_i\right)\right\},
\end{equation}

\begin{equation}
   \textit{F}_2\left(S_b^{-}, S_c^{+}\right) = \left\{\left(f_b\left(\widetilde{x}_i\right) \neq y_i\right) \cap \left(f_c\left(\widetilde{x}_i\right) = y_i\right)\right\}.
\end{equation}

The symbol $\cap$ represents the intersection. Filter functions filter out samples that are inconsistent between sub-model b and sub-model c. These samples are close to the decision boundary of the ensemble and need to be given greater weight during the training of model a. We redefine the adversarial risk as:

\begin{equation}
    \begin{aligned}
    R\left(\theta_a^{1}\right):= & \left[I\left(f_b\left(\widetilde{x}_i\right)=y_i\right) \cap I\left(f_c\left(\widetilde{x}_i\right) \neq y_i\right)\right] \\
     & + \left[I\left(f_b\left(\widetilde{x}_i\right) \neq y_i\right) \cap I\left(f_c\left(\widetilde{x}_i\right)=y_i\right)\right]. \\
    \end{aligned}
\end{equation}

There is minimal adversarial risk for both sub-models to correctly classify the samples and high adversarial risk for both sub-models to incorrectly classify the samples. We select and mark these samples as:

\begin{equation}
\textit{F}_3\left(S_b^{+}, S_c^{+}\right) = \left\{\left(f_b\left(\widetilde{x}_i\right) = y_i\right) \cap \left(f_c\left(\widetilde{x}_i\right) = y_i\right)\right\},
\end{equation}

\begin{equation}
\textit{F}_4\left(S_b^{-}, S_c^{-}\right) = \left\{\left(f_b\left(\widetilde{x}_i\right) \neq y_i\right) \cap \left(f_c\left(\widetilde{x}_i\right) \neq y_i\right)\right\}.
\end{equation}

These samples are far from the ensemble's decision boundary, and it is difficult to change the classification of the ensemble by single model, so we only need to use the standard adversarial risk:
\begin{equation}
\begin{aligned}
R\left(\theta_a^{2}\right):= & \left[I\left(f_b\left(\widetilde{x}_i\right)=y_i\right) \cap I\left(f_c\left(\widetilde{x}_i\right) = y_i\right)\right] \\
 & + \left[I\left(f_b\left(\widetilde{x}_i\right) \neq y_i\right) \cap I\left(f_c\left(\widetilde{x}_i\right) \neq y_i\right)\right]. \\
\end{aligned}
\end{equation}

We combine the different adversarial risks and train sub-model that minimizes the following risks:

\begin{equation}
\begin{aligned}
\min_\theta &\, R(\theta_a) 
:= \frac{1}{n} \left( \sum_{\textit{F}_1+\textit{F}_2} R(\theta_a^{1}) + \sum_{\textit{F}_3+\textit{F}_4} R(\theta_a^{2}) \right).
\end{aligned}
\end{equation}

We optimize separately samples that are consistent and inconsistent between model b and model c during the training of model a. Similarly, during the training of model b, we distinguish samples where model a and model b have consistent or inconsistent classifications. When training model c, we differentiate the samples where model a and model b have consistent or inconsistent classifications.

\subsection{Collaborative Ensemble Adversarial Training}

 According to \cite{b15}, distance metrics can provide more accurate evaluation and prediction. When training model a, we introduce the distance regularization:

\begin{equation}
\mathcal{L}_{adv} = \left|f_a\left(\widetilde{x}_i\right)-f_a\left(x_i\right)\right|^2.
\end{equation}

\begin{figure}[t]
    \centering
    \includegraphics[width=8cm, height=5cm]{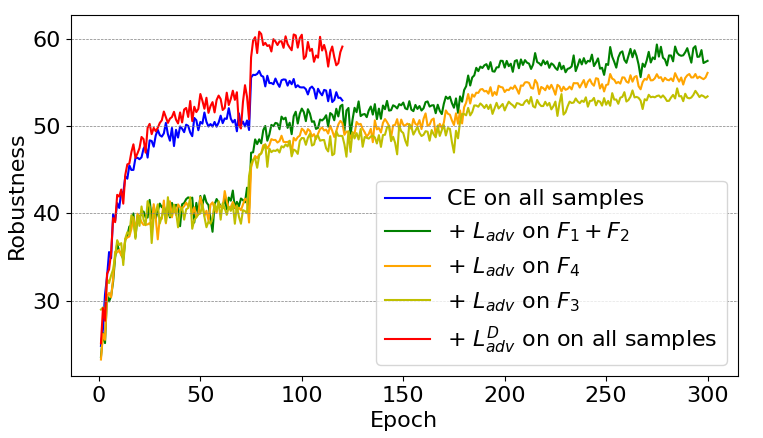}
    \caption{Comparison between standard adversarial training and introducing distance regularization on adversarial samples with consistent and  inconsistent classification between sub-models.}
    \label{fig:CE+L_adv}
\end{figure}

For the sample set $\textit{F}_1 \cup \textit{F}_2$, we apply only the cross-entropy(CE) loss. However, for the sample set $\textit{F}_3 \cup \textit{F}_4$, in addition to the cross-entropy loss, we also incorporate the distance loss $\mathcal{L}_{adv}$ during the training of sub-model a. As shown in Figure \ref{fig:CE+L_adv}, although the robustness has improved, the optimization process slows down. We analyze that optimizing 0-1 loss is difficult because it only exists when one sub-model correctly classifies the sample while another does not. When both sub-models make consistent classification, 0-1 loss disappears. The discontinuity of the 0-1 loss function makes gradients ineffective in providing meaningful optimization directions, making the optimization process slow. To solve this problem, we propose a continuous calibrating distance regularization to replace 0-1 loss.

We add distance loss $\mathcal{L}_{adv}$ to sample set $\textit{F}_3$ and $\textit{F}_4$ respectively, and find that the robustness does not improve. Therefore, we can conclude that special attention to sample set $\textit{F}_1 \cup \textit{F}_2$ enhances robustness. We can solve the problem of discontinuity by adding distance loss $\mathcal{L}_{adv}$ for all samples. However, it assigns the same attention to all samples, and cannot quantify the degree of probability disparities. Although reweighting techniques have been explored in model training\cite{b22}\cite{b23}, the idea in EAT has not been explored, and our work fills this gap. We define the prediction probability of model on real class of the sample as $h_m\left(\widetilde{x}_i\right) = p_m(\widetilde{x}_i, y_i)$, where $y_i$ represents the real class of sample. To measure the prediction disparities between sub-models on real class of the sample, we introduce a soft decision strategy which calculate absolute value between $h_b(\widetilde{x}_i)$ and $h_c(\widetilde{x}_i)$. To avoid discontinuity in loss function optimization and amplify the disparities, we define the exponential disparities:

\begin{equation}
e^{D} =  e^{\mu\left|h_b\left(\widetilde{x}_i\right)-h_c\left(\widetilde{x}_i\right)\right|}
\label{eq:prediction distance},
\end{equation}
where $h_b(\widetilde{x}_i)$ and $h_c(\widetilde{x}_i)$ are the confidence level predicted by sub-models b and c. If both can recognize correctly or incorrectly, then the prediction disparities will be very small, and the value will be close to constant 1. Otherwise, when one sub-model identifies correctly and one sub-model identifies incorrectly, the prediction disparities will be larger and the value will be amplified. Our proposed $e^{D}$ is multiplied by $\mathcal{L}_{adv}$ to obtain calibrating distance regularization, which is specifically designed for ensemble adversarial training:

\begin{equation}
\begin{aligned}
\mathcal{L}_{adv}^{D} =  e^{D} * \mathcal{L}_{adv}.
\label{eq:adv_L}
\end{aligned}
\end{equation}

It is well known that adversarial training improves robustness while inevitably decreasing accuracy. Subsequently, to explore the trade-off between accuracy and robustness, we attempt to solve the problem through the mutual cooperation between sub-models in ensemble adversarial training. In order to achieve the trade off between accuracy and robustness, we also differentiate clean samples where there are prediction disparities between the sub-models:

\begin{equation}
\mathcal{L}_{nat} = |f_a\left({x}_i\right)-y_i|^2,
\end{equation}

\begin{equation}
\begin{aligned}
\mathcal{L}_{nat}^{D} =  e^{\lambda\left|h_b\left(x_i\right)-h_c\left(x_i\right)\right|} * \mathcal{L}_{nat}
\label{eq:nat_L},
\end{aligned}
\end{equation}
where $y_i$ is the true label, and the $\lambda$ is used to amplify the prediction disparities of clean samples. Our proposed method consists of two parts: calibrating distance regularization for clean samples and adversarial samples based on predicted disparities. Adding this regularization to cross entropy loss for adversarial training, the training loss is:

\begin{equation}
\mathcal{L}_{total} = \mathcal{L}_{ce} + \lambda * \mathcal{L}_{nat}^{D} + \mu * \mathcal{L}_{adv}^{D}.
\end{equation}

Sub-model not only addresses its own adversarial vulnerabilities but also puts more concentrations on the samples based on prediction disparities. Instead of training independently, sub-models can cooperate and coordinate with each other.

\section{EXPERIMENT}

\subsection{Experimental Details}
\noindent{\bf Training Setup.} All the experimental results are completed on an NVIDIA 3090 GPU. We follow the experimental setup of previous paper. For the methods of ADP\cite{b6}, GAL\cite{b7}, and TRS\cite{b9}, sub-models share the Adam optimizer and run 150 epochs. The initial learning rate is set to 0.001 and decays tenfold at the 90th and 120th iterations. DVERGE\cite{b8}, LAFED\cite{b14}, FASTEN\cite{b11} and our method settings are the same. Sub-model uses a separate SGD optimizer and runs 120 epochs with momentum set to 0.9 and initial learning rate set to 0.01, decaying tenfold at the 75th and 95th iterations.

\noindent{\bf Whitebox robustness evaluation.}
We consider the following adversarial attacks to measure the white box robustness:
20-step Projected Gradient Descent(PGD) \cite{b2} with epsilon 0.031 and step size 0.007; 
20-step Momentum Iterative Method (MIM) \cite{b17} with epsilon 0.031 and step size 0.007;
Carlini \& Wanger Attack (CW) \cite{b18} with PGD optimization for 20 steps with epsilon 0.031 and step size 0.007; 
Auto-Attack (AA) \cite{b19}, consisting of FAB \cite{b20}, Square Attack \cite{b21} and two variants of PGD, is a combination of multiple attack methods.

\begin{figure}[htbp]
    \centering
    \includegraphics[width=11cm, height=6cm]{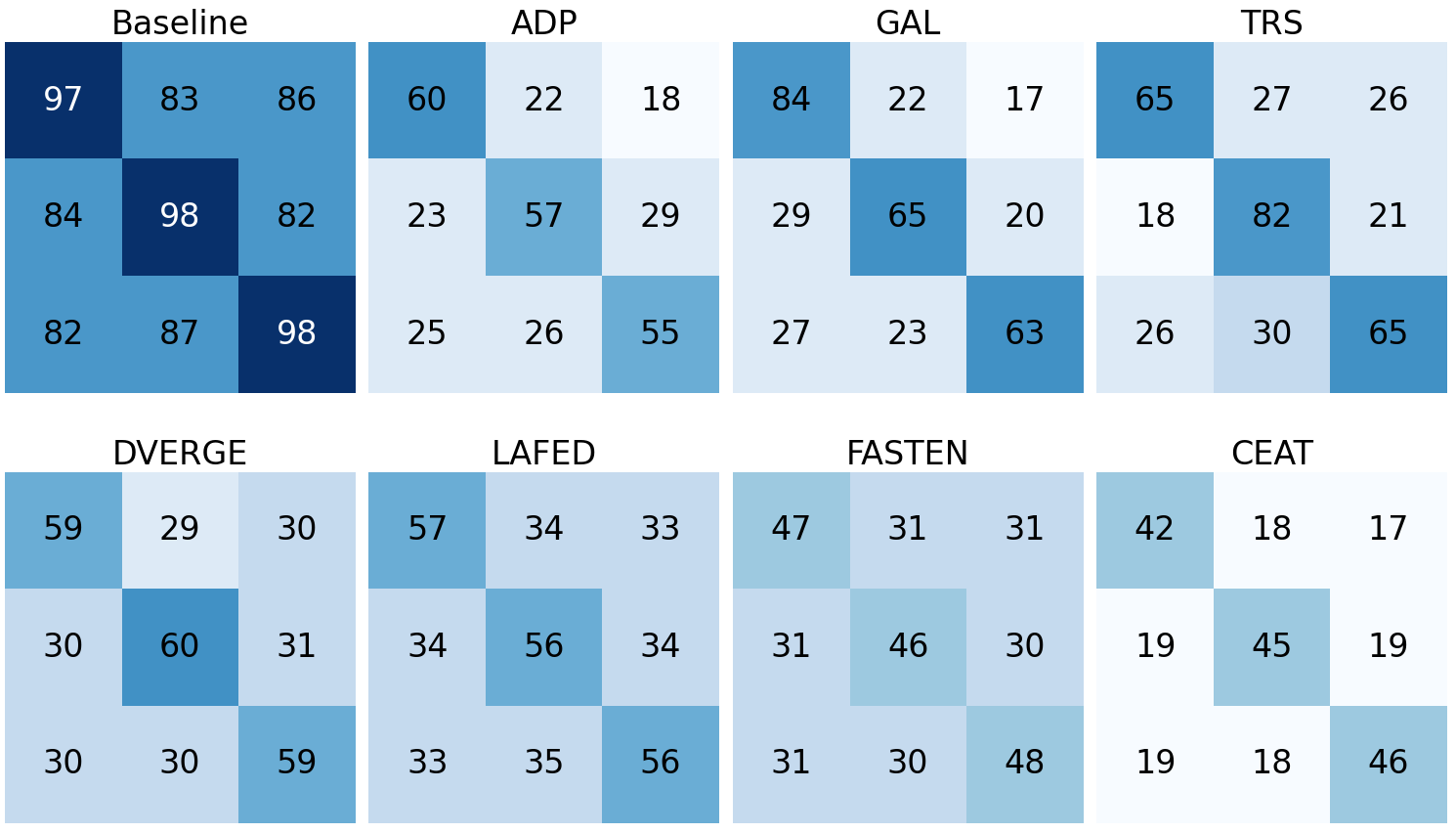}
    \caption{The success rate (\%) of adversarial attacks generated by each sub-model on itself and other sub-models. The lower the value, the better the defense of sub-models and the lower the transferability of the attack.}
    \label{fig:转移矩阵}
\end{figure}

\subsection{Comparison of transferability between sub-models}
Methods such as ADP\cite{b6}, GAL\cite{b7}, TRS\cite{b9} adopt optimization of $\mathcal{L}_{\mathrm{EAT}}$, however, sub-models share one optimizer. The inability of sub-models to update parameters based on their own losses limits the expressive power of individual models.  Moreover, adjusting optimization parameters may affect all sub-models, resulting in inevitable conflicts. Methods such as DVERGE\cite{b8}, LAFED\cite{b14}, and FASTEN\cite{b11} adopt optimization paradigm of $\mathcal{L}_{\mathrm{AT}}$, with independent optimizer optimizing for its own vulnerabilities. However, the ensemble vulnerabilities ignored by this optimization result in suboptimal defense capabilities against ensemble attacks. As shown in Figure \ref{fig:转移矩阵}, we test the success rate of adversarial sample attacks generated by each sub-model against itself and other sub-models, testing its own defense capabilities, and assessing its attack transferability to other sub-models. CEAT adopts $\mathcal{L}_{\mathrm{EAT}}$, with independent optimizer, achieving the lowest success rate for attacks, which means the defense effect is the best.

\begin{table*}[h]
\caption{Comparison with different methods on CIFAR-10 and CIFAR-100. NAT is the accuracy of classifying clean samples. PGD20, MIM, CW and AA are the robustness against four types of attacks. (5,1) and (1,5) are the values of parameters $\lambda$ and $\mu$.}
\renewcommand{\arraystretch}{1.5}  
\setlength{\tabcolsep}{3pt} 
\centering
\scriptsize  
\begin{tabular}{c|ccccc|ccccc}
\hline  \cline{1-11}
\multicolumn{1}{c|}{} & \multicolumn{5}{c|}{CIFAR-10}   & \multicolumn{5}{c}{CIFAR-100}  

\\ 
\hline  
Method & NAT    & PGD20  & MIM    & CW     & AA   & NAT    & PGD20  & MIM    & CW     & AA 
\\
\hline  \cline{1-11}
ADP    & 84.19  & 45.98  & 48.92  & 43.33  & 36.05 
& \underline{60.01}   & 21.18   & 22.48    & 19.39    & 15.81 

\\
GAL  & 79.64  & 42.56  & 44.12  & 39.82  & 34.70  
& 57.53   & 21.91   & 23.19    & 19.93   & 15.96 

\\
TRS  & 80.38  & 45.74  & 46.93  & 42.69  & 39.06  
& 53.98   & 23.66   & 24.27    & 20.72    & 18.11 

\\
\hline  \cline{1-11}
DVERGE & 85.38 & 48.72 & 50.48 & 47.73 & 44.27 
& 54.67   & 22.80   & 23.70    & 21.63    & 18.88

\\
LAFED  & \underline{86.94}  & 49.24  & 51.05  & 49.11  & 46.04 
& 59.27   & 25.12   & 26.31    & 23.91    & 21.53

\\
FASTEN & 81.21 & \underline{51.44}  & \underline{52.35}  & \underline{49.90}  & \underline{47.71} & 53.87   & \underline{26.06}   & \underline{26.65}    & \underline{24.33}    & \underline{22.32}

\\
\hline   \cline{1-11}

CEAT(5,1)   & \textbf{89.14}  & 52.07  & 54.14  & 51.06  & 46.74
& \textbf{63.90}   & 29.17 & 30.52 & 27.92 & 24.16

\\
CEAT(1,5)   & 85.74  & \textbf{59.25}  & \textbf{59.97}  & \textbf{54.58}  & \textbf{51.75}
& 59.90  & \textbf{32.18} & \textbf{33.45} & \textbf{30.62} & \textbf{27.72}

\\
\hline  \cline{1-11}
\end{tabular}
\label{tab:Experimental results of two datasets}
\end{table*}

\begin{table*}[h]
\caption{Experimental results of combining our proposed method as a module with previous methods on CIFAR-10. $\textit{L}^{D}$ is the calibrating distance regularization loss for natural samples and adversarial samples.}
\renewcommand{\arraystretch}{1.5}
\setlength{\tabcolsep}{6pt} 
\centering 
\scriptsize
\begin{tabular}{c|ccccc}
\hline  \cline{1-6}
         & \multicolumn{5}{c}{CIFAR-10}                            \\
\hline
Method         & NAT         & PGD20       & MIM         & CW          & AA          \\
\hline  \cline{1-6}
ADP (+$\textit{L}^{D}$)    & 84.19/83.60 & 45.98/\textbf{48.81} & 48.92/\textbf{50.37} & 43.33/\textbf{46.22} & 36.05/\textbf{42.97} \\
GAL (+$\textit{L}^{D}$)    & 79.64/82.42 & 42.56/\textbf{49.97} & 44.12/\textbf{51.11} & 39.82/\textbf{46.50} & 34.70/\textbf{44.15} \\
TRS (+$\textit{L}^{D}$)    & 80.38/79.26 & 45.70/\textbf{52.51} & 46.93/\textbf{53.88} & 42.69/\textbf{47.50} & 39.06/\textbf{45.74} \\
\hline  \cline{1-6}
DVERGE (+$\textit{L}^{D}$)   & 85.38/85.64 & 48.72/\textbf{50.47} & 50.48/\textbf{51.70} & 47.73/\textbf{50.07} & 44.27/\textbf{47.26} \\
LAFED (+$\textit{L}^{D}$)  & 86.94/85.67 & 49.24/\textbf{51.72} & 51.05/\textbf{53.09} & 49.11/\textbf{50.87} & 46.04/\textbf{48.40} \\
FASTEN (+$\textit{L}^{D}$) & 81.21/81.48 & 51.44/\textbf{52.57} & 52.35/\textbf{53.75} & 49.90/\textbf{50.40} & 47.71/\textbf{48.57} \\
\cline{2-2}
\hline \cline{1-6}
\end{tabular}
\label{tab:module}
\end{table*}

\subsection{Experimental Results}

We test the accuracy and robustness of several methods separately. Table \ref{tab:Experimental results of two datasets} shows that our proposed method achieves the excellent robustness performance improvement on datasets CIFAR-10 and CIFAR-100. Even compared to state-of-the-art methods, CEAT can achieve the best robustness in weak attack and strong attack methods. We can adjust the regularization parameters to achieve a trade-off between accuracy and robustness. To achieve higher accuracy, a larger prediction difference amplification coefficient for clean samples is preferred. If we want to achieve higher robustness,  a larger coefficient for adversarial samples is beneficial. For example, when the values of $\lambda$ and $\mu$ are 5 and 1, the accuracy of prediction for clean samples can reach 89.14\text{\%}; When the values of $\lambda$ and $\mu$ are 1 and 5, the robustness reaches the highest when defending against adversarial attacks. We also explore whether the method proposed in this paper can be integrated as a module with previous ensemble methods. Experimental results in Table \ref{tab:module} demonstrate that calibration distance regularization can serve as an independent loss that can be combined with previous methods and it can further enhance the robustness of the ensemble.

To validate the generalization, we not only use three identical ResNet-20 models for adversarial training but also employ three different models, including ResNet-20, ResNet-26, and ResNet-32, to enhance model learning diversity. Different model architectures can learn distinct features of the samples, avoiding co-adaptation overfitting while fully utilizing unique strengths of different model. As shown in Table \ref{tab:Extended experiments}, the accuracy is slightly lower than the best-performing FASTEN, but the robustness against adversarial attacks reaches optimal performance. We also extend our experiments to four sub-models. During the ensemble of four sub-models, the samples where two sub-models disagree in their predictions are assigned higher weights during the training of the other two sub-models, and these samples receive stronger regularization compared to the integration of three sub-models. The experimental results, as shown in Table \ref{tab:Extended experiments}, demonstrate an improvement in robustness; however, as the number of sub-models increases, computational costs and training times also significantly increase. In many research and applications, the ensemble of three sub-models has become a common choice.

\begin{table*}[h]
\caption{Experimental results of ensemble of three different sub-models and four identical sub-models.}
\renewcommand{\arraystretch}{1.5}  
\setlength{\tabcolsep}{3pt} 
\centering
\scriptsize  
\begin{tabular}{c|ccccc|ccccc}
\hline  \cline{1-11}
\multicolumn{1}{c|}{sub-models} & \multicolumn{5}{c|}{resnet20 resnet26 resnet32}   & \multicolumn{5}{c}{resnet20 resnet20 resnet20 resnet20}

\\ 
\hline  
Method & NAT    & PGD20  & MIM    & CW    & AA   & NAT   & PGD20  & MIM    & CW     & AA 
\\
\hline  \cline{1-11}
ADP  & 82.90	& 47.61	& 48.99	& 42.86	& 38.77 
& 84.56	& 45.63	& 47.16	& 41.55	& 37.33 

\\
GAL  & 79.38 & 51.89 & 52.90 & 49.35 & 46.53 
& 79.38	& 51.89	& 52.90	& 49.35	& 46.53

\\
TRS  &79.96	&44.40	&45.69	&41.39	&37.07 
&79.96	&44.40	&45.69	&41.39	&37.07

\\
\hline  \cline{1-11}
DVERGE &\underline{84.78}	&\underline{51.75}	&\underline{53.30}	&50.64	&47.77
&84.29	&50.97	&52.59	&50.68	&47.32 

\\
LAFED  &84.50	&50.94	&52.33	&49.22	&46.84
&\underline{84.85}	&51.50	&52.65	&50.06	&47.51

\\
FASTEN  &82.35	&51.67	&52.86	&\underline{50.69}	&\underline{48.27}
&82.65	&\underline{51.78}	&\underline{52.72}	&\underline{50.94}	&\underline{48.51}

\\
\hline   \cline{1-11}

CEAT   &\textbf{84.87}	&\textbf{58.16}	&\textbf{58.89}	&\textbf{55.25}	&\textbf{52.77}
&\textbf{85.14}	&\textbf{59.08}	&\textbf{59.93}	&\textbf{55.77}	&\textbf{53.16}

\\
\hline  \cline{1-11}
\end{tabular}
\label{tab:Extended experiments}
\end{table*}

\subsection{Ablation study}

\begin{figure}[htbp]
    \centering
    \begin{minipage}{0.48\textwidth}
        \centering
        \includegraphics[width=6cm, height=4.5cm]{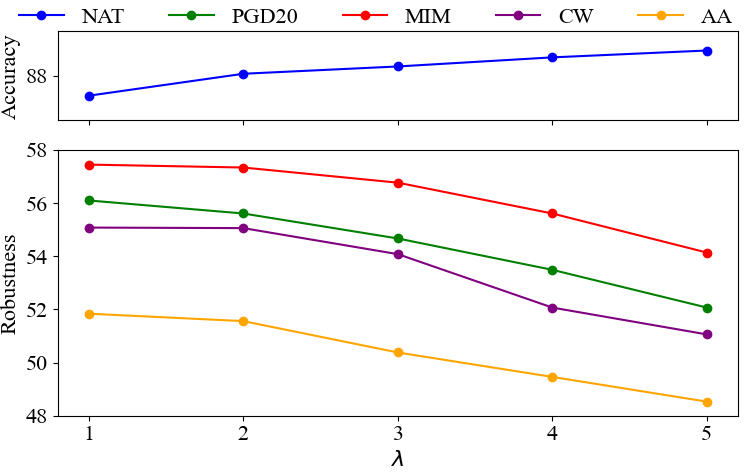} 
        \caption{Sensitivity to parameter $\lambda$.}
        \label{fig:Sensitivity to parameter1}
    \end{minipage}
    \hfill  
    \begin{minipage}{0.48\textwidth} 
        \centering
        \includegraphics[width=6cm, height=4.5cm]{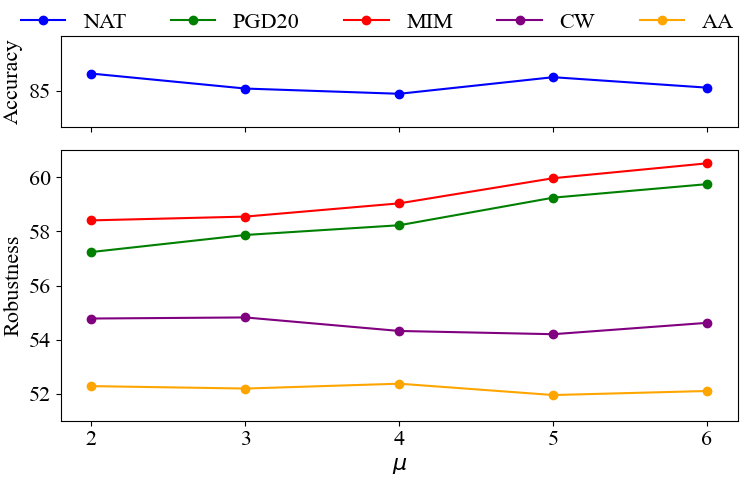} 
        \caption{Sensitivity to parameter $\mu$.}
        \label{fig:Sensitivity to parameter2}
    \end{minipage}
\end{figure}

The experiments are conducted on CIFAR-10 under three resnet-20 models. We fix the value of $\mu=1$ and take values of 1, 2, 3, 4, 5 for $\lambda$. We fix the value of $\lambda=1$ and take values of 2, 3, 4, 5, 6 for $\mu$, as shown in Figure \ref{fig:Sensitivity to parameter1} and \ref{fig:Sensitivity to parameter2}. The results show that our proposed method achieves stability and robustness at different coefficients. As the $\lambda$ amplifies, the accuracy increases, while the robustness decreases. As the $\mu$ amplifies, the robustness to weak attacks (PGD20 and MIM) increases, while the robustness to strong attacks (CW and AA) slightly decreases. Considering that there are no decreases in accuracy and an increase in robustness, we choose $\lambda=1, \mu=5$ for further ablation analyzes.

We conduct ablation experiments to investigate the effect of each component added to the loss function. As shown in 
Table \ref{tab:ablation study}, if only distance regularization is added, the robustness is improved by 2\text{\%} to 3\text{\%}, and the accuracy is reduced from 85.51\text{\%} to 83.11\text{\%}. If the prediction disparities is taken into account for training, the accuracy can be maintained at 83.98\text{\%}, which still decreases, but the robustness is improved to 60.48\text{\%} and 60.93\text{\%}. Considering the trade-off between accuracy and robustness, the distance regularization $\mathcal{L}_{\textit{nat}}$ of clean samples is introduced to maintain the accuracy of 85.74\text{\%}, which remains basically unchanged, while the robustness performance reaches 59.25\text{\%} and 60.21\text{\%}. The comparison of item 2 and item 3, item 4 and item 5 of the experimental results shows that the exponential amplification factor $e^{D}$ plays an important role in improving the robustness of the ensemble. Calibration distance regularization is introduced for clean samples to achieve the trade-off between accuracy and robustness. The comparison of experimental results between items 1 and 4 and items 1 and 5 shows that the robustness is significantly improved without reducing the accuracy.

\begin{table}[t]
\caption{Ablation study for each component of loss on CIFAR-10.}
\renewcommand{\arraystretch}{1.5}
\scriptsize
\begin{center}
\begin{tabular}{c@{\quad}|c@{\quad}c@{\quad}c@{\quad}|c@{\quad}c@{\quad}c@{\quad}}
\hline    \cline{1-7}
  & \( e^{D} \) & $\mathcal{L}_{\textit{adv}}$ & $\mathcal{L}_{\textit{nat}}$ & NAT   & PGD20  & MIM  \\
\hline    \cline{1-7}
1 &   &   &   & 85.51 & 56.36  & 57.40  \\
2 &   & $\vcenter{\hbox{$\sqrt{}$}}$  &  & 83.11  & 59.74   & 59.69  \\
3 & $\vcenter{\hbox{$\sqrt{}$}}$ & $\vcenter{\hbox{$\sqrt{}$}}$ &   & 83.98  & 60.48  & 60.93 \\
4 &   & $\vcenter{\hbox{$\sqrt{}$}}$ & $\vcenter{\hbox{$\sqrt{}$}}$ & 85.32 & 58.43  & 58.86  \\
5 & $\vcenter{\hbox{$\sqrt{}$}}$ & $\vcenter{\hbox{$\sqrt{}$}}$ & $\vcenter{\hbox{$\sqrt{}$}}$ & 85.74 & 59.25  & 60.21  \\
\hline   \cline{1-7}
\end{tabular}
\label{tab:ablation study}
\end{center}
\end{table}

\subsection{Black box attacks and Training time}

\begin{figure}[h]
    \centering
    \includegraphics[width=6cm, height=4cm]{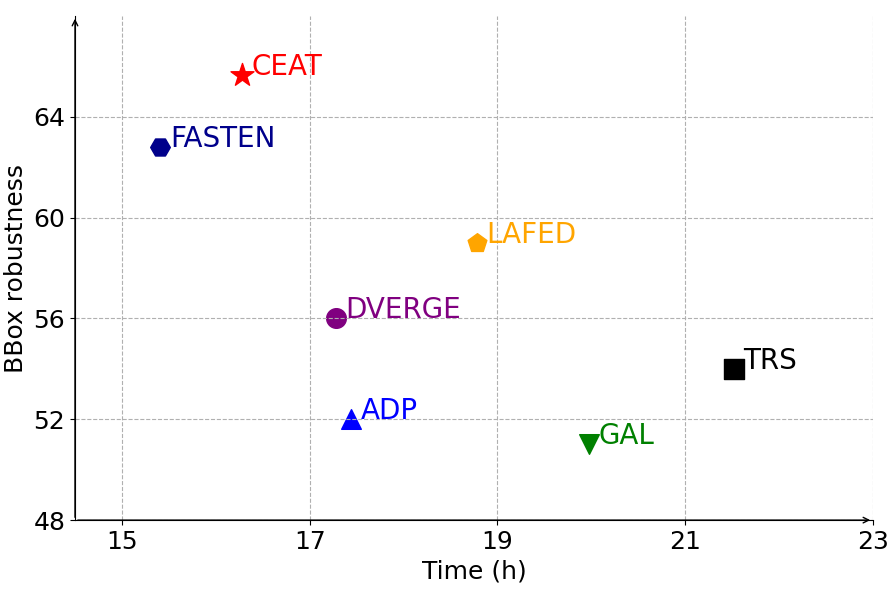}
    \caption{Black box robustness and Training time.}
    \label{fig:BBox and Time}
\end{figure}

Black box attacks cannot access the structure and parameters of the models. They use the transferability of the adversarial samples to attack by testing the proxy model. As shown in Figure \ref{fig:BBox and Time}, CEAT achieves the best robustness against black box attacks. Many ensemble methods have limited the application due to the high time cost involved in data augmentation and complicated regularization. CEAT removes data augmentation and uses a straightforward calibrating distance regularization, significantly reducing training time. While it is only marginally slower than FASTEN, its robustness is considerably superior. In contrast, other methods not only take longer time than CEAT but also exhibit much lower robustness.

\section{Conclusion}

In this paper, we introduce a novel and effective Collaborative Ensemble Adversarial Training (CEAT), in which sub-models are no longer trained independently but collaborated with each other. Experiments illustrate that our method can improve the robustness of the ensemble model on different datasets and models. Ablation experiments explicitly validate the effectiveness of each component in our method. In the future, we plan to integrate the fast adversarial training strategy into our method, further enhancing the computational speed in ensemble adversarial training.

\section*{\acknowledgmentsname} 
This work is supported in part by the Hunan Province Key Research and Development Program (Grant No. 2025QK3004), the NUDT Foundation (Grant Nos. 25-ZZCX-JDZ-39 and ZK24-27), the Changsha Excellent Young Innovators Program (Grant No. kq2209001), and the Hunan Excellent Young Scientists Fund (Grant No. 2025JJ40066). The authors also appreciate the helpful discussions with colleagues in the research group.

\end{document}